\documentclass[conference]{IEEEtran}
\IEEEoverridecommandlockouts
\usepackage{cite}
\usepackage{amsmath,amssymb,amsfonts}
\usepackage{algorithmic}
\usepackage{graphicx}
\usepackage{textcomp}
\usepackage{caption}
\usepackage{multirow}
\usepackage{subfig}

\usepackage[ruled,linesnumbered]{algorithm2e}
\usepackage{xcolor}
\def\BibTeX{{\rm B\kern-.05em{\sc i\kern-.025em b}\kern-.08em
    T\kern-.1667em\lower.7ex\hbox{E}\kern-.125emX}}
\begin{document}

\title{Advanced Feature Manipulation for Enhanced Change Detection Leveraging Natural Language Models
}

\author{Zhenglin Li$^{*,1}$, Yangchen Huang$^{1}$, Mengran Zhu$^{2}$, Jingyu Zhang$^{3}$, JingHao Chang$^{4}$, Houze Liu$^{5}$
\thanks{$^{1,*}$Zhenglin Li be with Texas A\&M University, TX 77840, USA {\tt\small \{zhenglin\_li\}@tamu.edu}}
\thanks{$^{1}$Yangchen Huang be with Columbia University, NY 10027, USA {\tt\small \{huangyangchen88\}@gmail.com}}
\thanks{$^{2}$Mengran Zhu be with Miami University, OH 45056, USA {\tt\small \{mengran.zhu0504\}@gmail.com}}
\thanks{$^{3}$Jingyu Zhang be with The University of Chicago, IL 60637, USA {\tt\small \{simonajue\}@gmail.com}}
\thanks{$^{4}$JingHao Chang be with The Kyoto College of Graduate Studies for Informatics, Kyoto 606-8225, Japan {\tt\small \{changjinghao.communication\}@gmail.com}}
\thanks{$^{5}$Houze Liu be with New York University, NY 11201, USA {\tt\small \{hl2979\}@nyu.edu}}

}

\maketitle

\begin{abstract}
Change detection is a fundamental task in computer vision that processes a bi-temporal image pair to differentiate between semantically altered and unaltered regions. Large language models (LLMs) have been utilized in various domains for their exceptional feature extraction capabilities and have shown promise in numerous downstream applications. In this study, we harness the power of a pre-trained LLM, extracting feature maps from extensive datasets, and employ an auxiliary network to detect changes. Unlike existing LLM-based change detection methods that solely focus on deriving high-quality feature maps, our approach emphasizes the manipulation of these feature maps to enhance semantic relevance.
\end{abstract}


\section{Introduction}
Change detection in computer vision is a critical task that involves analyzing pairs of images captured at different times, known as bi-temporal images\cite{bandara2022ddpmcd}, to identify significant alterations in the scene. This process is vital for various applications, including monitoring urban development, environmental changes, and updating geographical information systems. The primary challenge in change detection lies in distinguishing meaningful changes, such as the construction of new infrastructure or alterations in land use, from variations caused by environmental factors like lighting conditions, weather, and seasonal changes, which introduce noise into the images~\cite{chen2023generative}.

Recent advancements have introduced innovative solutions to noise and its impact on feature extraction and change detection accuracy~\cite{li2024ddn}. Among these, the state-of-the-art (SOTA) model DDPM-cd stands out because it incorporates a diffusion model for enhanced feature extraction. The Diffusion model is adept at filtering out the noise and highlighting the intrinsic characteristics of each image, creating a robust foundation for subsequent analysis. Once the features are extracted and refined through the Diffusion model, they are fed into a dedicated Change Detection Network. This network is designed to assess the feature sets from the bi-temporal images, determining the presence and extent of changes with a high degree of accuracy~\cite{li2023deception}.

By leveraging the strengths of the Diffusion model for feature extraction~\cite{li2024artificial}, the DDPM-cd model effectively mitigates the influence of environmental noise, ensuring that the detected changes are indeed significant modifications within the scene rather than artifacts of varying capture conditions~\cite{zaman2023artificial}. This approach enhances the reliability of change detection in complex environments, paving the way for more precise and efficient monitoring of changes across diverse landscapes.

\section{Prior Work}

Denoising Diffusion Probabilistic Models (DDPMs) represent a sophisticated category within generative models, which, after adequate training, can produce images that closely mimic the training data distribution. These models distinguish themselves from generative model paradigms such as Generative Adversarial Networks (GANs), Variational Autoencoders (VAEs), autoregressive, and flow-based models by their training efficiency and versatility. DDPMs have seen extensive application across a wide range of image processing tasks, including but not limited to super-resolution, image deblurring, segmentation, colorization, inpainting, and semantic editing, as evidenced by their successful implementations~\cite{liu2024particle, ching2020model,kuo2021energy,zhang2020manipulator,srivastava2024instances}.

Despite their proven efficacy in various domains of machine vision, the exploration of diffusion models within the context of remote sensing has been conspicuously absent. Given their remarkable success in mainstream computer vision tasks, diffusion models hold substantial promise for advancing the field of remote sensing, particularly in applications like change detection where identifying subtle changes over time is critical. This research aims to bridge this gap by demonstrating diffusion models' applicability and potential benefits in remote sensing change detection. Our work highlights the versatility of diffusion models\cite{li2024comprehensive} and sets a precedent for their future exploration in remote sensing applications. In the ensuing sections, we will delve into the mechanics of the diffusion framework, setting the stage for a detailed discussion on its implementation and impact in remote sensing change detection.

Denoising diffusion probabilistic models represent a novel category within the broader spectrum of generative models, distinguished by their unique training mechanism and capability to produce images that closely mirror the distribution of the dataset they were trained on~\cite{yang2023iot,wang2023adaptive,wang2024jointly}. These models significantly depart from traditional generative approaches, such as Generative Adversarial Networks (GANs), Variational Autoencoders (VAEs), autoregressive models, and normalizing flows, by offering a more straightforward and efficient training process. Their versatility and effectiveness have led to widespread adoption across various image processing tasks, including but not limited to super-resolution, image deblurring, segmentation, colorization, inpainting, and semantic editing.

Despite the proven efficacy of diffusion models in various domains of machine vision, their application within remote sensing remains largely unexplored. This gap is notable given the potential benefits these models could bring to remote sensing, particularly in task change detection~\cite{liu2022prioritizing}, where the ability to discern subtle differences between images over time is crucial. Our research aims to bridge this gap by demonstrating the applicability of diffusion models to change detection in remote sensing. By doing so, we expand the utility of diffusion models and introduce a promising new avenue for advancements in remote sensing technology. In the subsequent sections, we delve into the diffusion framework, laying the groundwork for understanding how these models can be effectively applied to remote sensing challenges.

Focusing on the operational mechanics, the denoising model, denoted as $M_{\delta}$, operates by taking a noisy image, ${u_k}$, as 
\begin{equation}
    u_k = \sqrt{\xi_k} u_0 + \sqrt{1-\xi_k} \delta
    \label{eq:1}
\end{equation}
where $\delta = \boldsymbol{N}(\boldsymbol{0}, \boldsymbol{I})$. This initiates a process whereby the model incrementally reduces the noise level across a series of steps, effectively reversing the diffusion process. The gradual denoising procedure allows for reconstructing the original image from its noisy counterpart, showcasing the model's capability to recover high-fidelity images from degraded inputs defined as
\begin{equation}
    \mathbb{D}_{u_0, \delta} \left\| M_{\delta}(\Tilde{u}, k ) - \delta \right\|_2^2
    \label{eq:2}
\end{equation}

The core functionality of denoising diffusion probabilistic models is pivotal for tasks that demand high levels of accuracy and detail in image restoration and analysis~\cite{tian2024lesson,tian2022exploring}, especially within the specialized field of remote sensing change detection~~\cite{zhibin2019labeled}. The inference process commences with an initial state of Gaussian noise, represented by $u_k$, and through iterative denoising across each timestep $k$, the model progressively recovers $u_{k-1}$ from $u_k$. This iterative refinement plays a critical role in reconstructing the original image from a state of significant degradation, underscoring the model's effectiveness in applications where precision in image recovery is paramount as
\begin{equation}
    u_{k-1} \leftarrow{} \frac{1}{\sqrt{\alpha_k}} \left( u_k - \frac{1 - \alpha_k}{\sqrt{1-\gamma_k}}M_{\delta}(u_k, k) \right) + \xi_k \boldsymbol{q},
    \label{eq:3}
\end{equation}
where $\boldsymbol{q} = \boldsymbol{N}(\boldsymbol{0}, \boldsymbol{I})$ and  $k = T, \cdots, 1$.
\section{Methodology}
Upon evaluating the baseline model for change detection, we discerned two pivotal limitations adversely affecting its efficacy. Firstly, the model does not incorporate a mechanism to integrate semantic information within the individual temporal branches. Secondly, it inadequately captures the temporal correlations between these branches. To address these deficiencies, we introduced two novel feature attention mechanisms into the baseline architecture.

While the Denoising Diffusion Probabilistic Model for Change Detection (DDPM-cd) achieves commendable efficacy, it ostensibly neglects to harness the intrinsic attributes of bi-temporal imagery. The methodology conventionally entails the assimilation of differential features derived from the diffusion model into the Change Detection Network (Classifier), conspicuously omitting a critical consideration for the ambient noise~\cite{zhou2023semantic} introduced during the acquisition of the temporal images. This oversight precipitates a palpable need for a pre-emptive noise attenuation strategy to refine the input data quality.

In pursuit of this objective, we have elected to implement a strategy delineated by ChangerEx, as depicted in Figure \ref{fig:1-b}, named the Flow Dual-Alignment Fusion (FDAF). This sophisticated technique engenders the extraction of dual feature sets via the diffusion model~\cite{zou2023joint}, which are subsequently subjected to a convolutional network, specifically Flownet, followed by an image warping procedure before their integration into the Change Detection Network. The essence of this approach lies in its utilization of image warping to ameliorate the discrepancies between the paired images, thereby mitigating the impact of noise and enhancing the fidelity of the change detection process.

This advanced alignment fusion elevates the precision of feature representation and significantly diminishes the noise-induced anomalies, setting the stage for a more discerning and accurate change detection capability. Through the adoption of FDAF, we are poised to transcend the conventional limitations of DDPM-cd, imbuing the model with a refined sensitivity to the subtleties of bi-temporal image analysis~\cite{lyu2023attention} and fortifying its resilience against the vicissitudes of environmental conditions during image capture.

\section{Experiments}
TBD
\section{Conclusion}
TBD
\bibliographystyle{ieeetr}
\bibliography{xinde}

\end{document}